\documentclass[letterpaper]{article} 
\usepackage{aaai2026}  
\usepackage{times}  
\usepackage{helvet}  
\usepackage{courier}  
\usepackage[hyphens]{url}  
\usepackage{graphicx} 
\usepackage{natbib}  
\usepackage{caption} 
\usepackage{algorithm}
\usepackage{algorithmic}

\usepackage{newfloat}
\usepackage{listings}
\DeclareCaptionStyle{ruled}{labelfont=normalfont,labelsep=colon,strut=off} 
\floatstyle{ruled}
\newfloat{listing}{tb}{lst}{}
\floatname{listing}{Listing}
\usepackage{listings}
\usepackage{verbatim}   
\usepackage{fancyvrb}   
\usepackage[most]{tcolorbox}
\usepackage{multirow}
\tcbuselibrary{breakable}  
\usepackage{xcolor}

\definecolor{lightgray}{rgb}{0.95, 0.95, 0.95}
\definecolor{darkgray}{rgb}{0.3, 0.3, 0.3}

\lstdefinestyle{json}{
    language=JSON,
    backgroundcolor=\color{lightgray},
    numbers=left,
    numberstyle=\tiny,
    stepnumber=1,
    numbersep=5pt,
    showstringspaces=false,
    breaklines=true,
    frame=single,
    basicstyle=\ttfamily\small,
    keywordstyle=\color{blue},
    stringstyle=\color{red},
    commentstyle=\color{green!60!black},
    title={\color{white}\textbf{LIWC Schema}}, 
    frame=tlrb,
    framesep=5pt,
    framerule=0.5pt,
    rulecolor=\color{gray!50!black},
    xleftmargin=5pt,
    xrightmargin=5pt
}

\newtcolorbox{jsonbox}[1][]{
    breakable,
    enhanced,
    colback=lightgray,
    colframe=gray!50!black,
    arc=0mm,
    boxrule=0.5pt,
    left=5pt,
    right=5pt,
    top=5pt,
    bottom=5pt,
    title={LIWC Schema},
    fonttitle=\bfseries,
    #1
}

\definecolor{systemprompt}{RGB}{220,230,242}
\definecolor{usermessage}{RGB}{242,242,242}
\definecolor{assistantmessage}{RGB}{236,246,236}

\lstdefinestyle{xmlcode}{
  basicstyle=\ttfamily\small,
  columns=flexible,
  keepspaces=true,
  breaklines=true,
  breakatwhitespace=false,
  showstringspaces=false,
  tabsize=2,
  escapeinside={(*@}{@*)},
  mathescape=true,
  literate={<}{{$<$}}1 {>}{{$>$}}1 {/}{{$/$}}1,
  breakindent=0pt,       
  xleftmargin=0pt,       
  xrightmargin=0pt,      
  postbreak=\mbox{\textcolor{gray}{$\hookrightarrow$}\space}  
}
\newtcolorbox{systempromptbox}{
  breakable,             
  enhanced,              
  colback=systemprompt,
  colframe=gray!50!black,
  arc=0mm,
  boxrule=0.5pt,
  left=5pt,
  right=5pt,
  top=5pt,
  bottom=5pt,
  title={Prompt},
  fonttitle=\bfseries,
}

\newtcolorbox{usermessagebox}{
  colback=usermessage,
  colframe=gray!50!black,
  arc=0mm,
  boxrule=0.5pt,
  left=5pt,
  right=5pt,
  top=5pt,
  bottom=5pt,
  title={User},
  fonttitle=\bfseries
}

\newtcolorbox{assistantmessagebox}{
  colback=assistantmessage,
  colframe=gray!50!black,
  arc=0mm,
  boxrule=0.5pt,
  left=5pt,
  right=5pt,
  top=5pt,
  bottom=5pt,
  title={Assistant},
  fonttitle=\bfseries
}
\usepackage{booktabs}

\title{PILOT: Steering Synthetic Data Generation with\\Psychological \& Linguistic Output Targeting}
\author{%
  Caitlin Cisar, Emily Sheffield, Joshua Drake, Alden Harrell,\\Subramanian Chidambaram,  Nikita Nangia,  Vinayak Arannil, Alex Williams
}
\affiliations{
    Amazon Web Services\\

  \texttt{\{caicisar,emilshef,jedx,jalden,subbuc,nangia,varannil,acwio\}@amazon.com}
}

\begin{document}

\maketitle

\begin{abstract}
  Generative AI applications commonly leverage user personas as a steering mechanism for synthetic data generation, but reliance on natural language representations forces models to make unintended inferences about which attributes to emphasize, limiting precise control over outputs. We introduce \textit{PILOT} (Psychological and Linguistic Output Targeting), a two-phase framework for steering large language models with structured psycholinguistic profiles. In Phase 1, PILOT translates natural language persona descriptions into multidimensional profiles with normalized scores across linguistic and psychological dimensions. In Phase 2, these profiles guide generation along measurable axes of variation. We evaluate PILOT across three state-of-the-art LLMs (Mistral Large 2, Deepseek-R1, LLaMA 3.3 70B) using 25 synthetic personas under three conditions: \textit{Natural-language Persona Steering} (NPS), \textit{Schema-Based Steering} (SBS), and \textit{Hybrid Persona-Schema Steering} (HPS). Results demonstrate that schema-based approaches significantly reduce artificial-sounding persona repetition while improving output coherence, with silhouette scores increasing from 0.098 to 0.237 and topic purity from 0.773 to 0.957. Our analysis reveals a fundamental trade-off: SBS produces more concise outputs with higher topical consistency, while NPS offers greater lexical diversity but reduced predictability. HPS achieves a balance between these extremes, maintaining output variety while preserving structural consistency. Expert linguistic evaluation confirms that PILOT maintains high response quality across all conditions, with no statistically significant differences between steering approaches. These findings establish PILOT as an effective framework for interpretable and controllable persona-based generation, bridging the gap between structured user modeling and nuanced linguistic expression in LLMs.

\end{abstract}

\section{Introduction}

Many downstream applications in Generative AI require large volumes of synthetic data that faithfully reflect a persona through the lens of their behavior \cite{noever2023ai}. However, existing methods for persona‐aligned data generation often struggle to provide precise, interpretable control over nuanced psycholinguistic traits \cite{jiang-etal-2024-personallm}. Fine‐tuning a model for each persona is costly and inflexible, while conventional prompt‐based steering often yields only coarse shifts in tone or style and can lead to overly explicit or generic outputs \cite{lester-etal-2021-power,li-liang-2021-prefix}.

Recent advances in prompt engineering demonstrate that LLMs can be instructed to adopt high‐level role descriptions (e.g., \textit{``You are a friendly tutor''}). While these approaches enable language models to adopt general behaviors of a desired persona, they lack mechanisms for fine‐grained modulation of linguistic signals such as confidence, authenticity, or analytical thinking \cite{caron-srivastava-2022-personality,jiang-etal-2024-personallm}. Psycholinguistic frameworks like LIWC and SEANCE offer validated dimensions across affective, cognitive, and social processes, but have primarily served post‐hoc analysis rather than active generation control \cite{pennebaker2015development,mairesse-walker-2011-controlling}. This gap between stylistic analysis frameworks and controllable generation limits our ability to produce outputs with quantitatively measurable stylistic properties while satisfying content requirements.

This paper introduces \textit{Psychological and Linguistic Output Targeting (PILOT)}, a two‐phase framework that leverages psycholinguistic profiles to steer generation along interpretable dimensions. In \emph{Phase 1}, PILOT constructs continuous style profiles for target personas by aggregating normalized scores across summary variables (e.g., analytical thinking), linguistic dimensions (e.g., pronoun use), and psychological processes (e.g., insight, affect). In \emph{Phase 2}, these profiles are injected into prompt schemas to guide model outputs toward specific linguistic patterns without requiring model retraining.

We evaluate PILOT in an ablation‐style study across three state‐of‐the‐art LLMs (Mistral Large 2, DeepSeek-R1, LlaMa 3.3 70B) and 25 synthetic personas derived from PersonaHub \cite{ge2024scaling} under three conditions: \textit{Natural-language Persona Steering (NPS)}, \textit{Schema‐Based Steering (SBS)}, and \textit{Hybrid Persona‐Schema steering (HPS)}. Our analyses reveal a fundamental trade-off between consistency and diversity in persona-based generation \cite{tseng2024talespersonallmssurvey}. Schema-based approaches (SBS and HPS) significantly reduced artificial-sounding persona repetition across all models. A cluster analysis showed that schema-based prompts also yield more coherent and topically-aligned outputs with higher silhouette scores and topic purity.

While SBS produced more concise outputs with moderate lexical diversity, NPS generated longer but more repetitive content. Most significantly, HPS achieved a balance between these extremes, maintaining output variety while preserving structural consistency. Human evaluation confirmed that incorporating PILOT schemas maintained output quality across all dimensions (overall quality, helpfulness, content adherence, and naturalness), with no statistically significant differences between steering approaches. Collectively, these results suggest that PILOT's schema-based prompting can more effectively control LLM outputs without diminishment to response quality.


\section{Related Work}
\label{sec:related}
\subsection{Steering Language Models with Personas}
Steering large language models (LLMs) to adopt specific personas or communication styles is an active area of research in NLP. Prior work has explored multiple strategies for controllable text generation, such as direct prompt-based control, psycholinguistic style manipulation, and structured schema-driven guidance. We review relevant work and highlight how our approach advances the frontier of persona-aligned, psycholinguistically controllable generation.

\subsubsection{Natural Language Personas}
Prompt-based methods steer LLM behavior by prepending natural language instructions or role descriptions to the input. This includes persona prompts (e.g., \textit{``You are a sarcastic storyteller''}), which guide tone, perspective, and language use without altering model weights \cite{gu-etal-2023-personas,jiang-etal-2024-personallm}. Studies show that LLMs can adopt such personas with moderate consistency, sometimes improving performance in downstream tasks \cite{li-etal-2023-lexicons,caron-srivastava-2022-personality}. However, persona prompts can also degrade accuracy if misaligned with the task \cite{mao2024editing}. Despite these trade-offs, natural language persona prompting remains popular due to its accessibility. It requires no retraining and enables lightweight control over traits like sentiment or formality, though subtler attributes (e.g., neuroticism or humor) often prove less steerable \cite{jiang-etal-2024-personallm}. Strongly aligned models may resist persona shifts to preserve helpfulness or safety. While fine-tuning can enhance steerability \cite{vu-etal-2024-psychadapter}, it is resource-intensive. 

\subsection{Schema-Based Steering}

While prompt tuning and style adjustments guide \textit{how} the LLM writes, schema-based methods provide control over \textit{what} the model writes. Traditional persona conditioning often relies on simple textual profiles (e.g., “John is a teacher who loves hiking”) \cite{zhang-etal-2018-personalizing}. Persona-Chat and similar datasets demonstrated that including such static profiles improves relevance and consistency in dialogue generation \cite{zhang-etal-2018-personalizing,urbanek-etal-2019-learning}. However, these methods typically operate over shallow descriptors and may result in repetitive or generic content.

Schema-based approaches combat shallow conditioning by introducing richer representations of persona behavior, such as scripts or event structures \cite{dunbar1997grooming,li-etal-2022-flashback}. These schemas provide models with behavioral context, such as what a teacher persona typically does on weekdays, in support of producing diverse, yet consistent responses. For example, \cite{lawley2023persona} proposed bootstrapping structured schemas from textual personas and using them to ground dialogue. Their findings indicate that schema-aware prompting leads to more coherent and content-rich output.

\subsection{Psycholinguistic Text Analytics}
Humans with different personalities or emotional states tend to differ in their word usage. Neurotic individuals, for example, tend use words that align to more negative emotions, and extroverts tend to use first-person plural pronouns \cite{pennebaker2001linguistic}. Linguistic Inquiry and Word Count (LIWC) \cite{pennebaker2015development} is a standard framework that quantifies such features. LIWC maps words to dozens of categories reflecting psychological, emotional, and social dimensions (e.g., \textit{affect}, \textit{insight}, \textit{pronouns}, \textit{formality}) \cite{pennebaker2015development}. By analyzing text against its curated lexicon, LIWC yields a profile of linguistic indicators, such as the proportion of words that reflect positive emotions or cognitive processes \cite{pennebaker2015development}. These profiles have been shown to correlate strongly with author traits and mental states \cite{pennebaker2001linguistic}.

\begin{figure*}[!ht]
    \centering
    \includegraphics[width=0.8\textwidth]{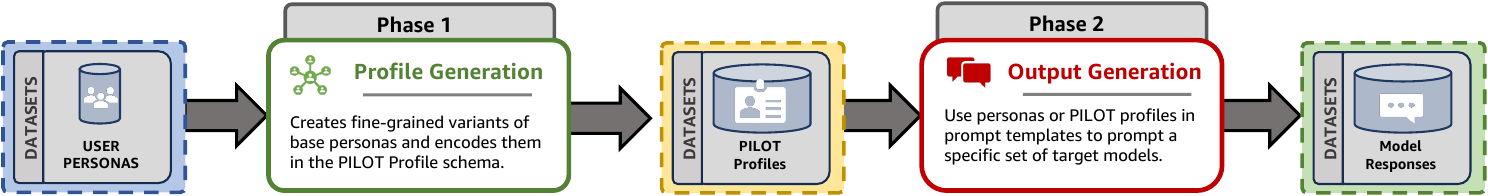}
    \caption{A flowchart illustrating the two-phase PILOT framework. In the first phase, we use an LLM to generate profiles in the form of schemas. In the second phase, these structured profiles are used to steer model generations by injecting the profiles into the model prompts.}
    \label{fig:pilot}
\end{figure*}

Some early systems demonstrated the feasibility of controlling stylistic expression through psycholinguistic features. PERSONAGE, for instance, varied utterances along personality axes using parametric linguistic features (e.g., verbosity, hedge words, self-references) derived from psychological findings \cite{mairesse-walker-2007-personage,mairesse-walker-2011-controlling}. Subsequent approaches attempted to control single dimensions such as politeness \cite{madaan-etal-2020-politeness} or sentiment using constrained decoding or reinforcement learning \cite{li-etal-2023-lexicons}. These tools, namely LIWC \cite{pennebaker2015development}, have historically been useful tools for conducting large-scale text analyses and have become increasingly used as tools for assessing the qualities of model-generated content \cite{jiang-etal-2024-personallm}. While this application demonstrates LIWC’s value for evaluation, using it as a proactive control signal for generation remains underexplored.



\subsection{Contribution}
Our work advances persona-based synthetic data generation by introducing a novel schema-based prompting technique that encodes personas as structured psycholinguistic profiles. Building on prior research, we fuse schema-driven conditioning with LIWC-based prompt controls to create the PILOT framework. This approach bridges the gap between abstract persona descriptions and concrete linguistic features, providing fine-grained control over generation without model retraining. We demonstrate that structured content grounding (via schemas) and psychologically-aligned style control (via PILOT) function as complementary mechanisms for persona alignment, enhancing both the consistency and naturalness of generated outputs. Together, they provide both depth and fidelity in controlled text generation, enabling more precise steering of language models along interpretable psychological and linguistic dimensions.


\section{Psychological and Linguistic Output Targeting}
\label{sec:pilot}
We introduce PILOT (Psychological and Linguistic Output Targeting), a two-phase framework for steering large language model outputs through structured psycholinguistic profiles. PILOT addresses a key limitation in current persona-based generation: natural language persona descriptions provide high-level guidance but lack fine-grained control over specific linguistic dimensions. Drawing from frameworks like LIWC \cite{pennebaker2015development} and SEANCE \cite{crossley2017sentiment}, PILOT represents personas as multidimensional profiles that quantify linguistic features across stability tiers.

Unlike conventional persona prompting, which relies on models to implicitly infer linguistic patterns, PILOT explicitly encodes these patterns in a structured schema. This approach bridges computational linguistics and psychological text analysis, providing an interpretable mechanism for steering outputs along validated dimensions of linguistic variation. By separating stable traits from context-sensitive features, PILOT enables nuanced representations that maintain coherence while adapting to different contexts.

\subsection{Schema Design}
Psycholinguistic research has demonstrated that certain linguistic features remain relatively stable across contexts while others demonstrate greater contextual sensitivity \cite{pennebaker2015development,biber2009register}. The PILOT framework formalizes both enduring personality traits and context-dependent linguistic behaviors as a hierarchical schema of dimensions spread across three tiers of stability:
\begin{itemize}
    \item \textit{Stable Dimensions}: Comprise linguistic features that remain highly consistent regardless of topic or communicative setting. Function words constitute the primary component of this tier, consistent with findings that these elements serve as reliable markers of cognitive style and personality traits \cite{pennebaker2001linguistic}.
    \item \textit{Semi-Stable Dimensions}: Encompass linguistic patterns that show moderate contextual adaptation while maintaining recognizable individual patterns. This category includes lexical diversity, referential cohesion, figurative language usage (metaphors, idioms), and sentence complexity metrics.
    \item \textit{Variable Dimensions} contain features that fluctuate substantially based on communicative context, audience, and subject matter. This comprehensive category includes pronominal distribution, parts of speech patterns, cognitive process markers, psychological drives, emotional valence, social behavior indicators, and numerous other contextually responsive dimensions.
\end{itemize}


This hierarchical organization offers several advantages for persona-based text generation. First, it provides a structured mechanism for representing multidimensional linguistic styles that can be systematically manipulated. Second, it enables more consistent persona alignment by encoding stable traits that persist across contexts. Third, it facilitates appropriate contextual adaptation by explicitly modeling which features should remain consistent and which should vary based on communicative demands.


\subsection{Framework \& Usage}
As shown in Figure \ref{fig:pilot}, PILOT implements persona-aligned generation through a two-phase operational framework that separates profile construction from text generation.

\subsubsection{Phase 1. Profile Generation}
In the first phase, PILOT translates a natural language persona description into a structured psycholinguistic profile aligned with our schema. This process employs a large language model (LLM) to analyze the persona description and representative text samples, mapping implicit linguistic characteristics to explicit dimensional values (0-100) across the schema's hierarchy.

Formally, the translation function $T$ maps a natural language description $P_{NL}$ to a structured profile $P_{PILOT} = {(d_1, v_1), (d_2, v_2), ..., (d_n, v_n)}$, where each dimension $d_i$ receives a normalized value $v_i$ based on its predicted prominence in the persona's linguistic behavior. This translation distills implicit stylistic expectations into explicit, manipulable parameters, enabling more precise and consistent persona representation.

\subsubsection{Phase 2. Output Generation}
In the output generation phase, PILOT uses the translated profile within a structured prompt template to guide an LLM's synthetic data generation process. 

PILOT's two-phase framework collectively offers several advantages over conventional persona-based generation approaches in terms. First, it enables precise quantification of linguistic features, allowing for systematic evaluation of profile adherence. Second, it provides an interpretable mechanism for steering generation along specific psycholinguistic dimensions. Third, it facilitates reuse of persona profiles across multiple generation tasks, enhancing consistency in persona-aligned applications.

\section{Experiment Design}
We designed an experiment to better understand PILOT's impact on the synthetic data generation process. We frame our experiment around a synthetic data generation task that focuses on seven content types that psycholinguistics frame as their targeted use-cases (e.g., personal letters, formal documents, social media posts) \cite{liwcapp2025}. We produce a total of 30 unique prompt instructions for each of the seven content types, yielding a total of 210 prompt instructions. For each instruction, we explore three persona-based steering strategies:
\begin{enumerate}
    \item \textit{Natural-language Persona Steering (NPS)}: Uses only the original natural language persona description (e.g., \textit{``an academic researcher"}) without structured guidance.
    \item \textit{Schema-Based Steering (SBS)}: Uses only the structured PILOT profile translated from the natural language description, providing the schema with dimensional values.
    \item \textit{Hybrid Persona-Schema Steering (HPS)}: Combines both the original persona description and its translated PILOT profile, integrating identity information with structured dimensional guidance.
\end{enumerate}

\subsection{Evaluation Dataset}
We sampled a diverse set of 250 natural language personas (and the texts associated with them) from PersonaHub \cite{ge2024scaling} and manually grouped them into 25 base personas such that each base persona contained 10 related sub-personas. For example, the base persona “academic researcher” includes sub-personas such as, "an academic researcher specializing in the history and impact of lecture series in universities, with a focus on the evolution of Whidden Lectures at McMaster University". We prompt Claude 3.5 Sonnet v2 to populate a PILOT profile for each synthetic output associated with each sampled sub-persona. The resulting vectors were averaged to create one reusable profile per base persona. All 25 PILOT profiles were validated to ensure the alignment and completeness of the schema.

\subsection{Evaluation Metrics}
We evaluate PILOT using a comprehensive set of metrics designed to assess three key aspects of persona-aligned generation: (1) steerability, (2) diversity, and (3) quality. These metrics enable systematic comparison across different prompting strategies and models.

\begin{itemize} \item \textit{Steerability Metrics:} We use cluster analysis to examine how different prompting strategies organize generated outputs. Higher silhouette scores \cite{rousseeuw1987silhouettes} indicate more distinct and cohesive output clusters, suggesting stronger steering influence. Cluster purity with respect to content requests \cite{manning2008introduction} reveals whether outputs are consistently aligned with intended topics. Smaller optimal cluster numbers with higher internal consistency suggest the prompting method exerts stronger structural control over output organization. Together, these metrics indicate whether schema-based prompts produce more predictable and consistently steered outputs than natural language prompts alone.

\item \textit{Diversity Metrics:} We quantify linguistic variation through n-gram diversity (i.e., ratio of unique n-grams to total n-grams), Type-Token Ratio (i.e., vocabulary richness measured as unique words divided by total words), and compression ratio (i.e., information density assessed via normalized text compression). Together, these metrics provide a multidimensional assessment of lexical richness, syntactic variety, and information content across generated outputs \cite{riaz2025metasynth}.

\item \textit{Content Quality Metrics:} We assess output quality through two complementary approaches: (1) automated analysis of persona repetition rates (i.e. percentage of responses containing explicit self-identification phrases), which indicates artificial-sounding persona adherence; and (2) expert human evaluation where trained linguists blindly rate 588 responses on 3-point Likert scales (1=low, 3=high) for overall quality, helpfulness, content adherence, and naturalness alongside qualitative feedback detailing the content's attributes that associate its authorship with a human or a model. For additional information, see Appendix C.
\end{itemize}

Statistical significance is assessed using Kruskal-Wallis tests with appropriate post-hoc analysis, and cross-model comparisons employ normalized metrics to account for model-specific baselines. All qualitative feedback (i.e., from human evaluation) was analyzed into high-level themes using Thematic Analysis \cite{clarke2014thematic}.

\subsection{Generation Protocol} 
We generate responses for each combination across the 25 personas, 3 steering method strategies, and 210 content requests for three state-of-the-art LLMs: a) LlaMa 3.3 70B Instruct, b) DeepSeek R1, c) Mistral Large 2. This generation procedure yielded a total of 47,250 responses (5,250 responses per steering strategy for each model)\footnote{All generation was facilitated by Amazon Bedrock via API.}.

\begin{figure}[!b]
    \centering
    \includegraphics[width=1.0\linewidth]{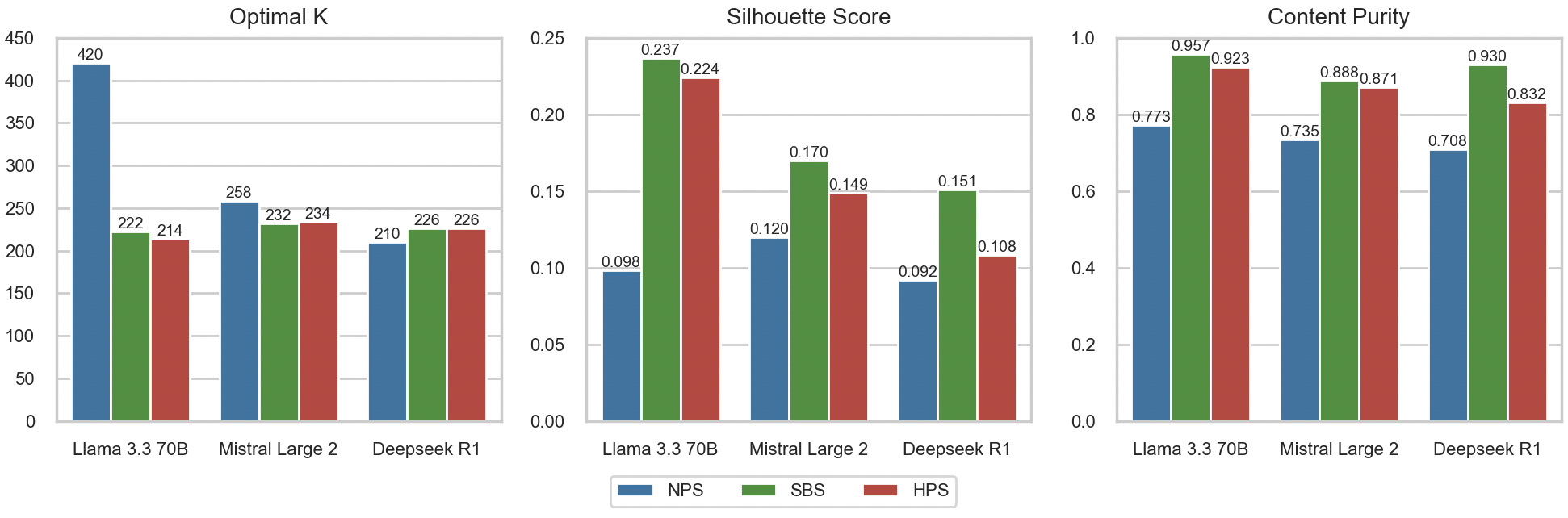}
    \caption{Steerability metrics for each steering strategy across LlaMa 3.3 70B, Mistral Large 2, and DeepSeek-R1.}
    \label{fig:steerability_bar_chart}
\end{figure}

\section{Results}
\label{sec:results}
\subsection{Steerability Analysis}

Figure \ref{fig:steerability_bar_chart} shows steerability metrics of all generated responses computed via clustering for all models and conditions. Silhouette scores, measuring cluster cohesion and separation, were highest under SBS conditions across all models, with LlaMa achieving 0.237 compared to 0.098 (NPS) and 0.224 (HPS). Cluster purity with respect to content request was consistently high under SBS conditions, with scores ranging from 0.888-0.957 compared to 0.708-0.773 under NPS and 0.832-0.923 under HPS. In contrast, persona purity scores were uniformly low across all conditions and models (0.100-0.182), with no condition achieving persona purity above 0.20. Optimal cluster numbers varied substantially across conditions, with NPS requiring the highest number of clusters (210-420) compared to SBS and HPS conditions (214-234). Average cluster sizes varied by condition, with NPS producing smaller clusters (2.38 responses) while SBS and HPS generated larger clusters (4.50 and 4.67 responses respectively). Within-cluster deviation increased progressively from NPS (2.45-3.00) to SBS (3.56-4.34) to HPS (4.38-7.47). We observed no statistically significant differences between the three targeted models.

\begin{figure}[!ht]
    \centering
    \includegraphics[width=1.0\linewidth]{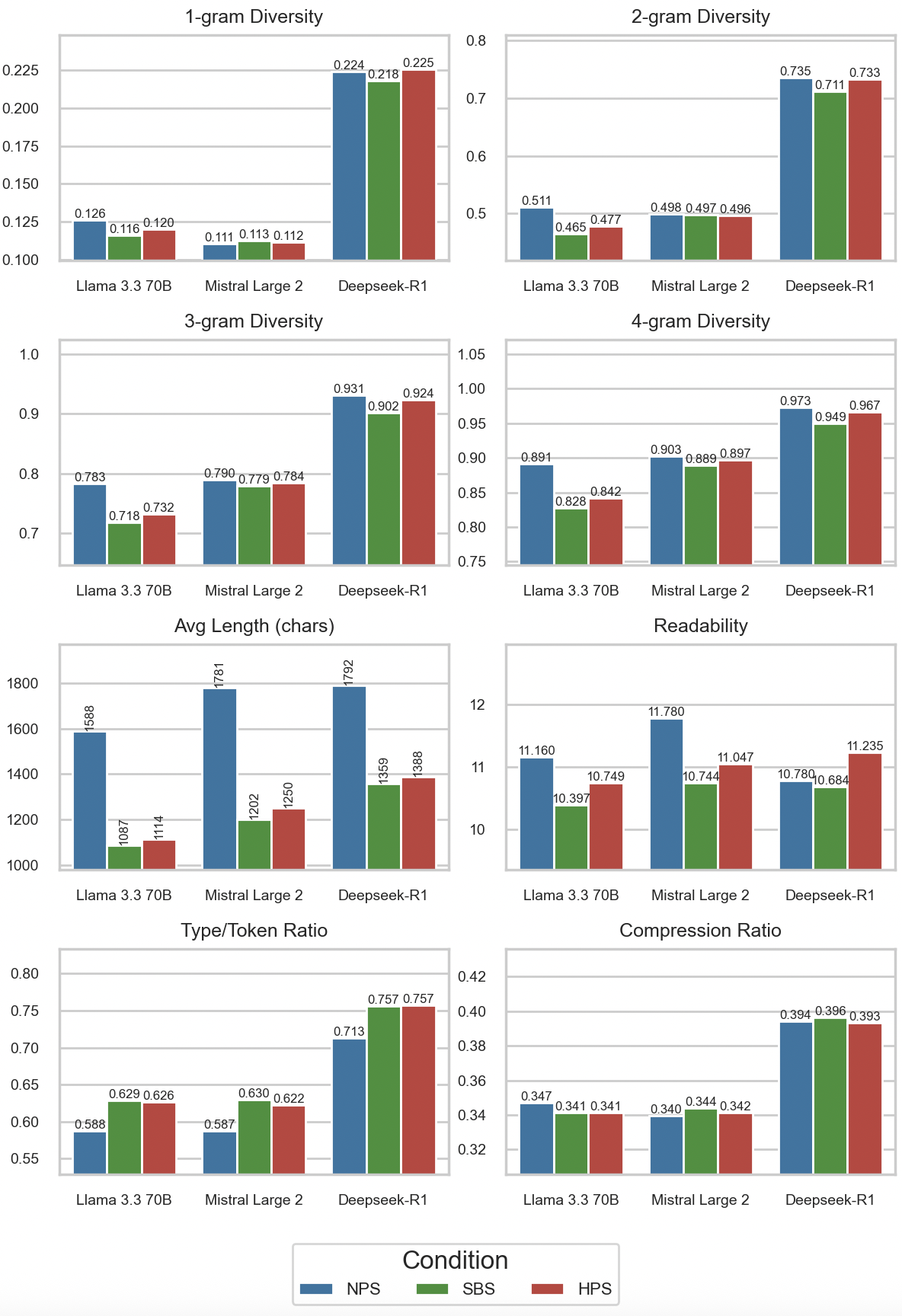}
    \caption{Diversity metrics for each steering strategy across LlaMa 3.3 70B, Mistral Large 2, and DeepSeek-R1.}
    \label{fig:diversity_bar_charts}
\end{figure}

\subsection{Diversity Analysis}
Figure \ref{fig:diversity_bar_charts} provides a visual summary of each diversity metric across all models and steering strategies. NPS consistently generated the longest outputs (1588-1792 characters) but demonstrated lower Type-Token Ratios across all models (0.59-0.71) compared to SBS (0.63-0.76) and HPS (0.62-0.76). SBS produced the most concise outputs (1087-1359 characters), while HPS fell between the two strategies. Higher-order n-gram diversity increased progressively from 1-gram to 4-gram across all strategies, with 4-gram diversity scores ranging from 0.83 to 0.97. SBS consistently showed the lowest n-gram diversity scores across all orders, while NPS and HPS showed similar patterns with slightly higher diversity scores. DeepSeek-R1 generated the longest outputs, maintaining the highest TTR scores (0.71-0.76) and substantially higher 1-gram repetition rates (0.22-0.23) compared to other models (0.11-0.13). Across all three strategies, readability scores remained stable (10.4-11.8 grade level), and compression ratios showed minimal variation (0.33-0.40).

\subsection{Quality Analysis}
\label{sec:qual}
\begin{table*}[ht]
\centering
\caption{Qualitative themes by steering strategy.}
\label{tab:qualitative_themes}
\begin{tabular}{llll}
\toprule
\textbf{Theme Category} & \textbf{NPS} & \textbf{SBS} & \textbf{HPS} \\
\midrule
\multirow{1}{*}{Self-Identification} & Explicit persona statements & Absence of persona statements & Moderate self-reference \\
\midrule
\multirow{1}{*}{Structural Patterns} & Template-like organization & Parallel sentence structures & Mixed structural patterns \\
\midrule
\multirow{2}{3cm}{Content Specificity} & Placeholder text & Keyword mirroring & Fewer placeholders \\
 & Generic content & Generic content & Generic content \\
\midrule
\multirow{2}{3cm}{Language Style} & Excessive formality & Impersonal tone & Sophisticated vocabulary \\
 & Repetitive phrasing & Formulaic language & Style inconsistencies \\
\midrule
\multirow{1}{*}{Perceived Authenticity} & Low naturalness & Moderate naturalness & Higher naturalness \\
\midrule
\multirow{2}{3cm}{Distinctive issues} & Self-referential framing & ``Vague, fake vibe"& Inconsistent perspective \\
 & Excessive structure & Fictitious details & Pronoun shifts \\
\midrule
\multirow{1}{*}{Criticisms} & High frequency & Moderate frequency & Low frequency \\
\bottomrule
\end{tabular}
\end{table*}

\subsection{Human Rating Data}
Our analysis of 567 responses across all four Likert metrics revealed generally positive evaluations across all experimental conditions. Overall, content adherence received the highest ratings (M=2.90, SD=0.33), followed by quality (M=2.47, SD=0.63), helpfulness (M=2.33, SD=0.72), and human-likeness (M=2.31, SD=0.76). This pattern suggests that while the generated outputs consistently adhered to requested content specifications, they were perceived as less natural and helpful. Two-way ANOVA revealed no significant interaction effects between prompting condition and model type for any of the four metrics, suggesting that the effectiveness of different prompting strategies did not depend on the choice of model. 

Strong positive correlations were observed between human-likeness and quality (r=0.68), human-likeness and helpfulness (r=0.70), and quality and helpfulness (r=0.71). Content adherence showed moderate positive correlations with the other metrics (r=0.25 to r=0.38). These correlations suggest that raters' perceptions of these attributes were interrelated, with outputs that seemed more human-like also being perceived as higher quality and more helpful.

\textit{Differences in Steering Strategies.} Statistical analysis revealed no significant differences between the three steering strategies across any of the four evaluation metrics. One-way ANOVA results showed no significant effects for human-likeness (F(2,564)=0.76, p=0.47), content adherence (F(2,564)=0.70, p=0.49), quality (F(2,564)=0.14, p=0.87), or helpfulness (F(2,564)=0.21, p=0.81). The absence of significant differences between steering strategies suggests that all three approaches produced outputs of comparable perceived quality, indicating that schema-based steering can achieve similar human evaluation scores to natural language prompting, despite using a more structured approach to persona representation.

\textit{Differences in LLMs.} We observed significant differences between language models. One-way ANOVA revealed significant effects of model type on human-likeness (F(2,564)=3.67, p=0.026), content adherence (F(2,564)=5.72, p=0.003), and helpfulness (F(2,564)=6.20, p=0.002), though not for quality (F(2,564)=1.75, p=0.17). Post-hoc Tukey HSD tests showed that LlaMa (M=2.42, SD=0.74) was rated significantly more human-like than Mistral (M=2.21, SD=0.77, p=0.019). For content adherence, DeepSeek (M=2.96, SD=0.20) significantly outperformed Mistral (M=2.85, SD=0.39, p=0.003). For helpfulness, both LlaMa (M=2.42, SD=0.68, p=0.003) and DeepSeek (M=2.38, SD=0.70, p=0.018) significantly outperformed Mistral (M=2.18, SD=0.77).

\textit{Differences in Content Type.} Content type had a substantial impact on perceived quality across all metrics. Content Type. Commercial writing and traditional book/short story formats received the highest overall ratings, while social media and correspondence formats received lower ratings. Social media content showed the lowest content adherence scores (M=2.59, SD=0.57), significantly lower than other content types that averaged above 2.90. Further analysis revealed significant condition effects within specific content types. For correspondence, quality ratings differed significantly by condition (F(2,78)=6.23, p=0.003). For social media content, we found significant differences by condition in content adherence (F(2,78)=6.46, p=0.003), quality (F(2,78)=3.68, p=0.030), and helpfulness (F(2,78)=3.16, p=0.048).

\subsubsection{Human Feedback Data}
Thematic analysis of qualitative feedback data from human evaluators (n=567) revealed distinctive patterns across steering strategies despite their statistical equivalence in ratings. Table \ref{tab:qualitative_themes} provides an overview of the themes observed across steering strategies.

\textit{Natural-language Persona Steering}
NPS outputs were characterized by explicit self-identification (e.g., ``As a researcher..."), with evaluators noting frequent use of persona role statements in 43.5\% of responses with LlaMa, 43.2\% with Mistral, and 13.4\% with DeepSeek. This condition exhibited template-like structures and repetitive sentence patterns in 38\% of cases, suggesting that natural language persona prompts lead models to emphasize persona identity through explicit framing. Placeholder text (e.g., ``[Professor's Name]") appeared in 27\% of NPS outputs, indicating recognition of needed specificity without sufficient context to provide it. Evaluators consistently identified excessive formality as inappropriate for many contexts in 35\% of NPS responses.

\textit{Schema-Based Steering}
SBS outputs demonstrated different characteristics, most notably the absence of explicit persona statements that were prevalent in NPS. Instead, evaluators described 31\% of these outputs as ``too impersonal" or lacking authentic personality. A distinctive feature was direct mirroring of schema keywords in generated content (identified in 24\% of responses), suggesting literal interpretation of the provided parameters. While still exhibiting formulaic structures in 29\% of cases, these manifested more as parallel sentence constructions rather than template-like organization. Evaluators noted the presence of obviously fictitious personal information in 18\% of outputs, contributing to what they described as a ``vague, fake vibe" in approximately 22\% of SBS responses.

\textit{Hybrid Persona-Schema Steering}
HPS outputs showed more nuanced characteristics, with style inconsistencies—particularly in formality levels and pronoun usage—being a distinctive feature in 25\% of responses. These outputs were noted for sophisticated vocabulary (e.g., ``labyrinth," ``mosaic," ``annals of film history") in 19\% of cases, sometimes to the point of appearing unnatural. Notably, HPS received fewer critical comments overall (23\% fewer than NPS and 17\% fewer than SBS) and was described as exhibiting fewer obvious AI indicators in evaluator comments, suggesting this approach may produce outputs perceived as more natural. However, inconsistency in personal perspective was a distinctive issue in 21\% of HPS responses, with shifts between first, second, and third person sometimes occurring within the same text.

\textit{Cross-Condition Patterns.} The most consistent indicators of AI-generated text across all conditions were: placeholder text (present in 27\% of NPS, 12\% of SBS, and 15\% of HPS outputs), generic content lacking specificity (identified in 38\%, 31\%, and 26\% of outputs respectively), inappropriate formality levels (35\%, 28\%, and 25\%), repetitive structures (38\%, 29\%, and 24\%), and template-like organization (40\%, 29\%, and 22\%). Content judged as more human-like typically included natural language variation, unique details, authentic-sounding dialogue, and contextually appropriate formality levels, with approximately 34\% of HPS outputs receiving positive comments about naturalness compared to 27\% for SBS and 24\% for NPS.

\section{Discussion}
\label{sec:conclusion}
Our study suggests that PILOT enhances persona-aligned generation through structured psycholinguistic profiles, as demonstrated in our ablation study across three LLMs (Mistral Large 2, Deepseek-R1, LlaMa 3.3 70B) comparing unstructured (NPS), structured (SBS), and hybrid (HPS) approaches. Although prior work relied on natural language personas \cite{gu-etal-2023-personas, jiang-etal-2024-personallm} or parameter-efficient tuning \cite{lester-etal-2021-power, li-liang-2021-prefix}, PILOT advances the field by explicitly encoding dimensions from frameworks such as LIWC \cite{pennebaker2015development, tausczik2010psychological}, providing explicit control over the generation procedure.

\subsection{Empirical Findings}

\subsubsection{Steerability and Coherence} Our cluster analysis demonstrates that schema-based approaches (i.e., SBS and HPS) achieved superior topical coherence compared to natural language prompting. While prior work \cite{samuel2024personagymevaluatingpersonaagents} previously observed that structured evaluation frameworks reveal more consistent patterns in persona-based generation than free-form assessments, our work takes this insight further by implementing structure directly into the generation process's instruction itself. Our findings indicate that PILOT's schema-based approach not only evaluates, but also actively shapes more reliable steering mechanisms than natural-language personas.

\subsubsection{Diversity-Consistency Trade-offs} Our diversity metrics reveal an important trade-off in persona-based generation that builds upon observations by \cite{tseng2024talespersonallmssurvey}, who identified the tension between consistency and diversity as a key challenge. Our contribution lies in demonstrating how this trade-off can be systematically managed: while SBS produced more concise outputs with moderate lexical diversity, NPS generated longer, but more repetitive content. Notably, we found that HPS achieved a balance between these extremes, suggesting a novel approach where combining schemas with natural language prompts can help maintain output variety while preserving structural consistency.

\subsubsection{Human Evaluation and Quality Assessment} Human evaluation further illuminates PILOT's contributions, showing that structured prompting maintains quantitative performance while enhancing qualitative engagement. The dramatic reduction in explicit self-identification phrases indicates that PILOT helps models avoid artificial-sounding persona repetition, a common issue identified by \cite{liu-etal-2024-evaluating-large} in persona-steered generation. Our work advances beyond this identification of the problem by providing a concrete mechanism to address it, demonstrating that structured psycholinguistic profiles can significantly reduce this unwanted behavior across different model architectures.

Thematic analysis of evaluator feedback reveals that each steering strategy produces distinctive linguistic patterns, with HPS outputs exhibiting more sophisticated vocabulary and fewer obvious AI indicators compared to other conditions. While \cite{Li2024HelloAL} demonstrated that personalized agents benefit from both explicit identity information and implicit stylistic guidance, our work operationalizes this insight through a concrete framework that systematically combines these elements. The reduction in critical comments for HPS outputs (23\% fewer than NPS and 17\% fewer than SBS) not only confirms their findings but extends them by showing how this hybrid approach can be implemented to produce outputs perceived as more natural.

\subsection{Relationship to Prior Work} Our work builds upon previous efforts to control linguistic style through psycholinguistic features. Early systems like PERSONAGE \cite{mairesse-walker-2007-personage, mairesse-walker-2011-controlling} demonstrated the feasibility of controlling stylistic expression through parametric linguistic features, but were limited to template-based generation systems. PILOT advances this approach by integrating modern LLMs with structured psycholinguistic profiles, enabling far more flexible and natural-sounding outputs while maintaining stylistic control. Unlike approaches that focus on single dimensions such as politeness \cite{madaan-etal-2020-politeness} or specific personality traits \cite{li-etal-2023-lexicons}, PILOT provides a comprehensive framework for multidimensional stylistic control that can be applied across diverse personas and communicative contexts.

\subsection{Limitations}
Our research has several limitations. First, our study focused on synthetic personas drawn from PersonaHub \cite{ge2024scaling}; real-world persona diversity (e.g., cultural backgrounds, language varieties, etc.) may present new challenges and require additional adaptation of the PILOT framework. Second, consistently low persona purity scores across all conditions suggest fundamental limitations in achieving fine-grained stylistic control through prompting alone. This indicates that while PILOT improves topical coherence, there remain challenges in achieving precise stylistic alignment with intended personas. Third, while our evaluation emphasizes stylistic and lexical metrics, downstream task performance (e.g., persuasiveness, engagement) remains to be assessed in real-world applications. PILOT's practical utility in the wild requires further investigation.

\subsection{Conclusion}
We introduced PILOT, a framework that advances controllable text generation by encoding psycholinguistic dimensions in structured profiles that guide LLM outputs. Our experiments demonstrate that schema-based approaches significantly enhance output coherence and topical alignment while maintaining response quality across multiple evaluation dimensions. The framework's hierarchical organization of linguistic features, ranging from stable function words to context-sensitive psychological markers, enables more nuanced stylistic control than traditional natural language prompting techniques. By bridging psycholinguistic theory with practical implementation, PILOT offers a principled method for steering language model outputs.

\bibliography{aaai2026}

\appendix
\section*{Appendix A: PILOT Schema}
The PILOT schema is a structured JSON‐based representation of psycholinguistic dimensions that are inspired by the LIWC framework \cite{pennebaker2015development}. The schema defines the set of stylistic and psychological features that aim to steer models in generating their responses. Each profile encodes continuous scores for summary variables, linguistic categories, psychological processes, and domain‐specific lexicons, enabling precise, multi‐dimensional control over an LLM’s output.

\begin{lstlisting}[caption={PILOT Schema}, label={lst:pilot_schema}]
"stability_hierarchy": {
  "stable": {
    "function_words": {}
  },
  "semi_stable": {
    "lexical_diversity": {
      "ttr": {}
    },
    "referential_cohesion": {},
    "figurative_language": {
      "metaphor": {},
      "idiom": {}
    },
    "sentence_complexity": {
      "average_sentence_length": {},
      "subordination": {},
      "embedding": {}
    }
  },
  "variable": {
    "pronouns": {
      "personal_pronouns": {
        "first_person_singular": {},
        "first_person_plural": {},
        "second_person": {},
        "third_person_singular": {},
        "third_person_plural": {}
      },
      "impersonal_pronouns": {}
    },
    "determiners": {
      "articles": {},
      "numbers": {}
    },
    "parts_of_speech": {
      "prepositions": {},
      "auxiliary_verbs": {},
      "adverbs": {},
      "conjunctions": {},
      "negations": {},
      "verbs": {},
      "adjectives": {}
    },
    "cognitive_processes": {
      "insight": {},
      "causation": {},
      "discrepancy": {},
      "tentative": {},
      "certainty": {},
      "memory": {}
    },
    "psychological_drives": {
      "affiliation": {},
      "achievement": {},
      "power": {}
    },
    "emotional_tone": {
      "positive_emotion": {},
      "negative_emotion": {
        "anxiety": {},
        "anger": {},
        "sadness": {}
      },
      "swear_words": {}
    },
    "social_behavior": {
      "prosocial": {},
      "politeness": {},
      "conflict": {},
      "moralization": {},
      "communication": {}
    },
    "social_references": {
      "family": {},
      "friends": {},
      "female_references": {},
      "male_references": {}
    },
    "abstractness": {
      "concrete_words": {},
      "abstract_words": {}
    },
    "hedging_language": {
      "basic_hedges": {},
      "epistemic_hedging": {},
      "modal_hedging": {},
      "approximation": {},
      "attribution_hedging": {},
      "politeness_hedging": {}
    },
    "discourse_markers": {},
    "cultural_references": {
      "politics": {},
      "ethnicity": {},
      "technology": {}
    },
    "lifestyle_domains": {
      "leisure": {},
      "home": {},
      "work": {},
      "money": {},
      "religion": {}
    },
    "physical_states": {
      "health": {
        "illness": {},
        "wellness": {},
        "mental_health": {},
        "substances": {}
      },
      "sexual": {},
      "food": {},
      "death": {}
    },
    "motivational_states": {
      "need": {},
      "want": {},
      "reward": {},
      "risk": {}
    },
    "perceptual_processes": {
      "visual": {},
      "auditory": {},
      "feeling": {},
      "motion": {},
      "space": {}
    },
    "conversational_features": {
      "informal_digital_language": {
        "slang": {},
        "internet_abbreviations": {}
      },
      "assent": {},
      "nonfluencies": {},
      "fillers": {}
    }
\end{lstlisting}

\section*{Appendix B: Prompts}

\subsection*{B.1 PILOT Profile Generation}
\begin{systempromptbox}{}
\begin{lstlisting}[style=xmlcode]
You are a sophisticated linguistic analysis tool that combines LIWC-style (Linguistic Inquiry and Word Count) profiling with higher-level discourse analysis. You analyze text across three stability levels: 

1. Stable dimensions (variation <=5%): Basic linguistic features like function word usage
2. Semi-stable dimensions (variation 10-15%): Complex linguistic patterns including:
  - Lexical diversity (vocabulary richness)
  - Referential cohesion (how entities are maintained through coreference)
  - Figurative language (metaphor and idiom usage)
  - Sentence complexity patterns
3. Variable dimensions (variation 20-30%): Context-dependent features like LIWC's traditional psychological, emotional, and topical categories

Your task:
1. Analyze the following input text
2. Return all metrics as percentages in the provided JSON schema
3. Use the JSON schema provided below.
4. Respond with ONLY the raw JSON object - do NOT include any explanations, markdown formatting, or introductory text.

    JSON schema:
    {schema}

    Text to analyze:
    {text}
    
Output: JSON object ONLY, following the schema with each value as a percentage (0-100).
\end{lstlisting}
\end{systempromptbox}

\subsection*{B.2 Response Generation}
\subsubsection*{B.2.1 Natural Language Prompt}
\begin{systempromptbox}{}
\begin{lstlisting}[style=xmlcode]
Pretend you are a(n) {persona}. {request}
\end{lstlisting}
\end{systempromptbox}

\subsubsection*{B.2.2 Schema Prompt}
\begin{systempromptbox}{}
\begin{lstlisting}[style=xmlcode]
Using the following profile as a linguistic style guide:
{profile}

Respond to this request: {request}.

Before you generate anything, think about what parameters from the profile are most important, explain your thought process, and then generate the response to the request. Provide your output in the following format: 
<explanation> explanation </explanation>

<response> response </response>

Now begin your analysis and response:
\end{lstlisting}
\end{systempromptbox}

\subsubsection*{B.2.3 Hybrid Prompt}
\begin{systempromptbox}{}
\begin{lstlisting}[style=xmlcode]
Pretend you are a(n) {persona}.

Using the following profile as a linguistic style guide:
{profile}

Respond to this request: {request}.

Before you generate anything, think about what parameters from the profile are most important, explain your thought process, and then generate the response to the request. Provide your output in the following format: 
<explanation> explanation </explanation>

<response> response </response>

Now begin your analysis and response:
\end{lstlisting}
\end{systempromptbox}
\subsubsection*{B.2.4 25 Personas}
The 25 natural language personas used for the NPS and HPS conditions are as follows: high school literature teacher, casual moviegoer, creative writer, architectural historian, mathematician, high school chemistry teacher, landscape architect, travel agent, graphic designer, computer scientist, software engineer, data scientist, AI researcher, wargaming enthusiast, academic researcher, military historian, film critic, high school history teacher, museum curator, film historian, film studies scholar, sports journalist, freelance journalist, civil engineer, researcher 

\subsubsection*{B.2.5 7 Content Types}
Content requests were generated in the following categories: personal writing, formal Writing, email correspondence, social media, commercial writing, entertainment (e.g. film or music), traditional book or story 

\subsection*{B.2.6 Commercial Writing Request}
\begin{tcolorbox}[colback=lightgray,colframe=gray!50!black,title=Request,fonttitle=\bfseries]
Write a product review (100–300 words) for a new wireless noise-canceling headset, focusing on performance, comfort, and value.
\end{tcolorbox}

\section*{Appendix C. Additional Details on Human Evaluation}
\label{app:human_eval_details}
To complement our quantitative findings, we conducted a qualitative evaluation of a sample of output responses using expert linguists. We collected 196 responses from each of three LLMs, resulting in a total evaluation corpus of 588 responses. We used identical persona, PILOT schemas, and content requests across all models. Our evaluation framework employed trained human linguists who assessed each response on four critical dimensions:
\begin{itemize}
\item Overall Quality: Aggregate perception of clarity, structure, grammatical correctness, and appropriate tone.
\item Helpfulness: Perceived effort required for the generated content to be considered useful.
\item Content Adherence: Perceived alignment with the original request's desired specifications.
\item Naturalness: Perceived similarity to human-authored text.
\end{itemize}
All ratings used a 3-point Likert scale (1=low, 3=high). Linguists were kept blind to both the source model and the prompting condition for each response. Additionally, linguists were asked to provide qualitative feedback on what made responses seem human or AI-generated, and to note any particularly distinctive aspects of each response.

\end{document}